\documentclass[conference]{IEEEtran}
\usepackage[spanish, es-tabla]{babel}
\IEEEoverridecommandlockouts
\usepackage{cite}
\usepackage{amsmath,amssymb,amsfonts}
\usepackage{algorithm}
\usepackage{algorithmic}
\usepackage{graphicx}
\usepackage{float}

\usepackage{textcomp}
\usepackage{xcolor}

\decimalpoint

\usepackage{balance}

\def\BibTeX{{\rm B\kern-.05em{\sc i\kern-.025em b}\kern-.08em
    T\kern-.1667em\lower.7ex\hbox{E}\kern-.125emX}}

\begin{document}

\title{Paralelización del submuestreo de nube de puntos para plataformas con memoria unificada}

\author{\IEEEauthorblockN{
Martin Nievas\IEEEauthorrefmark{1}, 
Claudio J. Paz\IEEEauthorrefmark{3} 
and Gastón R. Araguás\IEEEauthorrefmark{4}}

\IEEEauthorblockA{CIII - Centro de Investigación en Informática para la Ingeniería\\
Universidad Tecnológica Nacional 
Facultad Regional Córdoba, Argentina}


\IEEEauthorblockA{
\IEEEauthorrefmark{1}mnievas@frc.utn.edu.ar, \IEEEauthorrefmark{3}cpaz@frc.utn.edu.ar, \IEEEauthorrefmark{4}garaguas@frc.utn.edu.ar}
}


\maketitle

\begin{abstract}
La exploración de ambientes desconocidos mediante robots es una tarea que integra distintas áreas como localización por un lado y mapeo y planificación por otro.
En lo que respecta al mapeo, existen diversos métodos para representar los ambientes por los que puede transitar un robot, en dos y tres dimensiones.
Se pueden mencionar \textit{grilla de ocupación probabilística}, \textit{Octomap} y \textit{STVL} entre los más importantes los últimos años.
En la actualidad, las cámaras RGB-D son ampliamente utilizadas para generar una representación detallada del ambiente.
La misma presenta un gran volumen, el cual debe ser reducido para poder utilizarse en plataformas de recursos limitados de cómputo.

En este trabajo se presenta una implementación del método de diezmado de la nube de puntos capaz de ser ejecutado en una placa con memoria unificada.
El mismo consiste en reducir la nube de puntos de manera iterativa utilizando una subdivisión ordenada del espacio.
Se obtuvieron resultados para distintos tamaños de grillas, plataformas y escenarios tanto reales como simulados.
Los resultados indican que en sistemas embebidos es conveniente contar con arquitecturas que compartan memoria entre CPU y GPU para optimizar los procesos de paso de bloques de datos.\\

\end{abstract}


\section{Introducción}
En robótica, se conoce como exploración al proceso mediante el cual se busca aumentar la información respecto del entorno del robot para construir un modelo del ambiente que lo rodea.
Se aplican los algoritmos de exploración cuando se necesita conocer el estado de una locación a la que no pueden acceder personas o hay un riesgo latente o manifiesto en su ingreso. 
Por ejemplo en estructuras colapsadas o con posibilidad de colapso en busca de sobrevivientes.
En estas circunstancias los robots más utilizados son de dimensiones reducidas por su capacidad de atravesar aberturas pequeñas.
Debido a estos limitantes geométricos, la capacidad de cómputo de estos robots también se ve limitada.

Lo modelos generados por estos robots son útiles para poder navegar en estos ambientes, entendiéndose por navegar a la tarea que permite llevarlo de un punto a otro de manera segura evadiendo obstáculos.
Los modelos pueden ser métricos, donde se trata de representar o reproducir la geometría del entorno; o topológicos, donde se describe la relación espacial entre distintos ambientes.

Respecto de los métricos, durante años una de las formas más populares de representar el entorno fue mediante grilla de ocupación \cite{moravec_1985}.
Este método consiste en representar el ambiente bidimensional con un plano dividido en celdas formando una grilla.
Las celdas de la grilla son de tamaño regular y cada una contiene un valor que representa la probabilidad de estar ocupada basada en modelos probabilísticos de los sensores involucrados.
Derivaciones de este trabajo usando tres dimensiones con cubos llamados \textit{voxels} se presentaron en trabajos como \cite{moravec1996} y \cite{dryanovski2010} mostraron ser ineficientes respecto del uso de memoria para almacenar el mapa debido al gran tamaño de los mismos.

\begin{figure}[tp!]
\centering
  \includegraphics[width=0.45\textwidth]{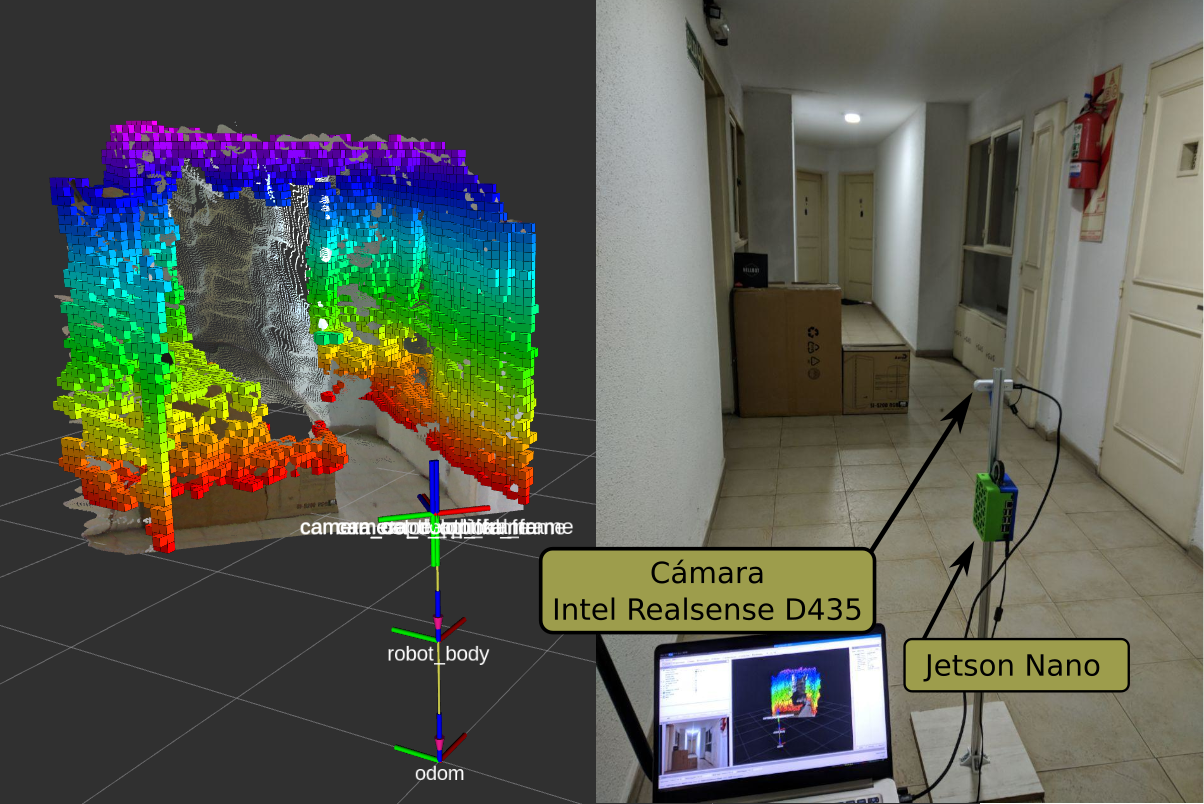}
  \caption{Montaje utilizado para las pruebas en un escenario real}
  \label{fig:pruebas_real}
\end{figure}

\IEEEpubidadjcol

De forma más eficiente, el \textit{framework} Octomap \cite{hornung2013} usa una estructura en forma de árbol con ocho nodos, donde cada nodo comienza con un voxel que es dividido sucesivamente en otros ocho hasta alcanzar la resolución deseada.
Esta estructura es llamada \textit{octrees} y fue propuesta por primera vez en \cite{doctor1981}.
El valor contenido en el voxel puede ser binario, un valor de probabilidad que puede estar basado en distintos criterios o una variante con una cota aplicada a una densidad de probabilidad.
Si bien el Octomap hace mejor uso de la memoria, el acceso a cada elemento tiene mayor costo computacional que la grilla de ocupación.

Para mejorar el acceso a cada elemento, en \cite{museth2013vdb} se propone una variante de árboles B+, usados generalmente en sistemas de archivos, llamada VDB por las siglas de \textit{Volume Dynamic B+tree}.
Con esta topología se puede modelar un espacio de índices tridimensional, \textit{virtualmente infinito} que permite el acceso rápido a información dispersa.
Adicionalmente, la implementación de VDB no impone restricciones de topología sobre los datos de entrada y admite patrones de acceso, inserción y eliminado aleatorio rápidos, en promedio $\mathcal{O}(1)$.
La implementación de esta estructura de datos es conocida como OpenVDB (OVDB) y se presenta en \cite{museth2019}.

Explotando las características de OVDB, en \cite{stvl2020steve} se presenta una biblioteca llamada STVL por las siglas de \textit{Spatio-Temporal Voxel Layer} donde se implementan una serie de \textit{buffers} que almacenan nubes de puntos provenientes de cámaras de profundidad u otras fuentes capaces de generar este tipo de datos.
Estas nubes de puntos se codifican usando OpenVDB logrando formar mapas tridimensionales de manera eficiente.
Debido a la resolución y cantidad de sensores, las nube suelen estar compuestas de millones de puntos por lo que es necesario realizar una \textit{compresión} diezmándola con algún criterio.
En \cite{stvl2020steve} esto es realizado mediante un \textit{filtro de voxelizado}, disponible en la biblioteca \textit{Point Could Library} \cite{Rusu_ICRA2011_PCL} el cual corre en CPU.

En este trabajo se presenta una implementación de STVL capaz de ser ejecutada en la GPU de la plataforma de desarrollo Jetson Nano de Nvidia, utilizando el arreglo de la Fig.~\ref{fig:pruebas_real}.
Esta plataforma fue elegida debido a su reducido tamaño y gran eficiencia energética, lo que posibilita su uso en pequeños robots ya sean voladores o terrestres.
Adicionalmente, esta plataforma tiene la CPU y la GPU en el mismo chip, por los que comparten físicamente la memoria.
Esto evita que se hagan copias redundantes de bloques de memoria, por ejemplo una imagen, entre el CPU y la GPU.
Este enfoque es conocido por NVIDIA como \textit{Zero-Copy}.
Específicamente se presenta una variante de \cite{bkedkowski2013general} implementada en CUDA para realizar la compresión de los puntos.
Esta compresión o filtrado se realiza aprovechando el acceso \textit{Zero-Copy} disponible en la plataforma Jetson Nano.
No hay, según el conocimiento de los autores, otra implementación que pueda ser ejecutada en esta plataforma.

El trabajo se organiza de la siguiente manera: En la Sección~II se describen las herramientas de \textit{software} y las plataformas de \textit{hardware} utilizadas en el trabajo, así como la propuesta de modificación al algoritmo para poder ser utilizado en la plataforma Jetson Nano. También se describen los escenario donde fue probada la implementación. En la Sección~III se muestran los resultados obtenidos en las distintas plataformas y escenarios, con tres tamaños de voxel. Las conclusiones del trabajo y futuras líneas de  investigación son enumeradas en la Sección~IV.

\section{Materiales y métodos}

Para evaluar el algoritmo propuesto se hicieron pruebas en un ambiente simulado y en un escenario real.
Ambos escenarios de similares características, estáticos y estructurados.
Además se comparó el tiempo de cómputo de la biblioteca original y la propuesta usando distintas plataformas.

\subsection{Herramientas de software utilizadas}
La implementación se realizó sobre el \textit{framework} ROS (siglas en inglés de Sistema Operativo para Robots) \cite{quigley2009ROS}, usando la versión de nombre clave Melodic Morenia.
Esto permite una rápida integración del método a evaluar con algoritmos existentes y ampliamente usados en robótica.
La mejora propuesta es una variante de método de filtrado de puntos de PCL~\cite{pcl} utilizado por la biblioteca STVL disponible en \cite{stvl}.
Para la etapa de simulación, la misma se realizó con el entorno Gazebo \cite{koenig2004design} versión 9 la cual puede descargarse de \cite{gazebo} e integrarse a ROS.

\subsection{Plataformas de hardware utilizadas}
La plataforma elegida para realizar las pruebas fue la Jetson Nano de NVIDIA.
La misma es una placa de pequeñas dimensiones, tan solo 70mm de largo y 45mm de ancho, lo que posibilita su utilización en robots de tamaño reducido.
Cuenta con una GPU de arquitectura Maxwell de 128 CUDA cores con los cuales puede correr hasta 2048 hilos y como procesador principal tiene un ARM Quad-core Cortex-A57.
Esta placa cuenta con una memoria de 4GB 64-bit LPDDR4 con un ancho de banda máximo teórico 25.6GB/s.
Con todo este \textit{hardware} puede alcanzar 472GFLOPS de rendimiento de cómputo en FP16, con 5-10W de consumo de energía.
Como se mencionó antes la CPU y la GPU se encuentran en el mismo encapsulado y comparten físicamente la misma memoria del sistema, admitiendo una forma limitada de \textit{memoria unificada}.
Dicha arquitectura permite comunicaciones CPU-GPU, que no son posibles en GPU discretas.
De esta forma, se evita la copia de cada punto y solamente es necesario pasar la dirección de memoria del mismo a las funciones de CUDA lo que se traduce en una reducción de tiempo y espacio utilizado.
Es importante destacar que esta implementación es en si misma una mejora sobre el algoritmo original presentado por \cite{bkedkowski2013general}, ya que originalmente estaba pensado para arquitecturas de GPU NVIDIA Fermi y Kepler las cuales son generaciones anteriores y no disponen de memoria unificada.

El algoritmo también fue evaluado en una PC de escritorio con procesador Ryzen 7 1700 y una GPU NVIDIA GTX 1660 Super.
La misma cuenta con 6GB de memoria GDDR6 de video, con un ancho de banda teórico máximo indicado por el fabricante de hasta 336GB/s.
Esta placa pertenece a la familia de micro arquitectura Turing la cual dispone de la tecnología de memoria unificada.
Como contrapartida, la memoria está separada del procesador de la CPU por lo cual la información tiene que ser copiada a través del bus PCI Express.

Para generar la nube de puntos en el escenario real se utilizó la cámara RGB-D Intel RealSense Depth Camera D435.
Esta cámara está formada por un par estéreo capaz de determinar la distancia al sensor de los puntos dentro de su campo de visión.
También dispone de un proyector infrarrojo, para mejorar la imagen 3D en paredes que no tengan características sobresalientes.

\subsection{Algoritmo propuesto}
En el algoritmo presentado en \cite{stvl2020steve}, las nubes de puntos provenientes de las cámaras RGB-D, son ingresadas en \textit{buffers} de observaciones, los cuales almacenan la posición relativa de la medición con respecto a un sistema de coordenadas globales.
Luego, estas mediciones pueden ser utilizadas con dos finalidades: operaciones de marcado en la cual se designa la celda como ocupada, u operaciones de liberación donde se marca la celda como vacía.

Debido a la implementación paralela del programa, existen \textit{buffers} extra destinado a cada una de las operaciones.
Las nubes de puntos recibidas, al estar en el sistema de referencia de la cámara, tienen que ser transformadas al sistema de coordenadas de globales.
Esta transformación de un sistema de referencias a otro es realizado mediante la biblioteca \texttt{tf2} \cite{foote2013}, la cual permite realizar las transformaciones entre los múltiples sistemas de referencia presentes en el robot, a lo largo del tiempo.

\begin{figure}[bt]
\centerline{\includegraphics[width=0.5\textwidth]{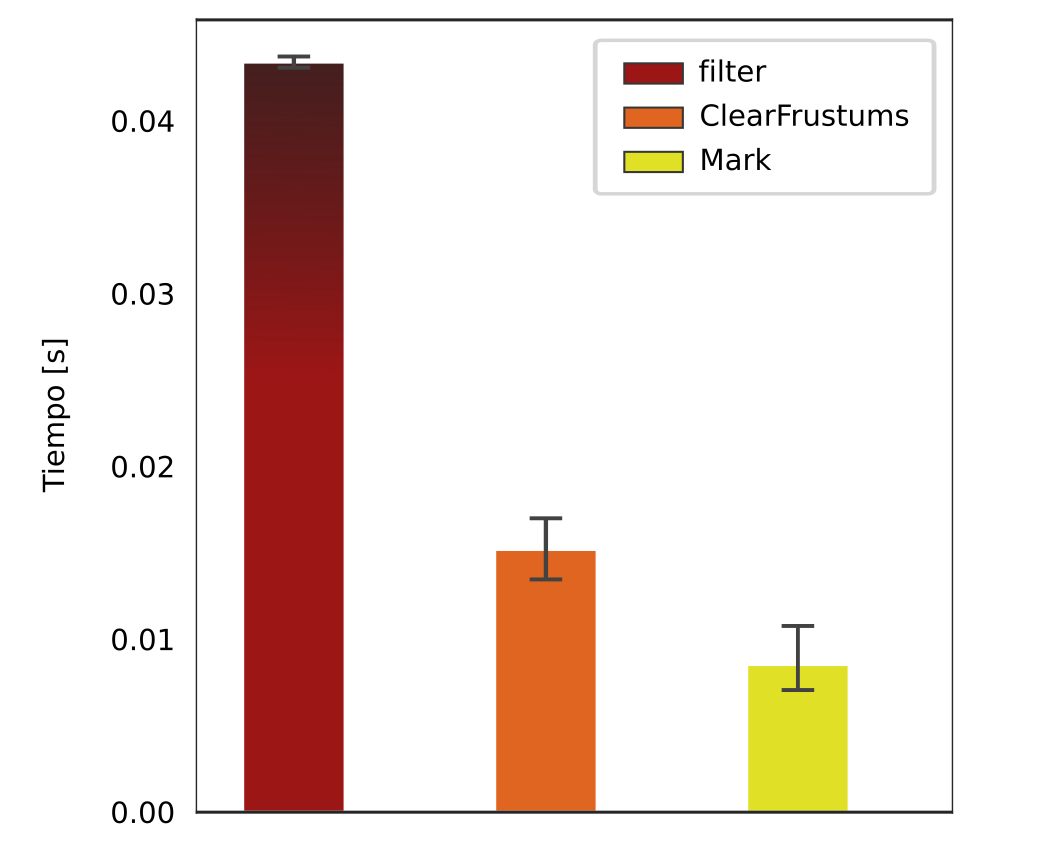}}
\caption{Tiempos de procesamiento de las partes de mayor carga computacional del algoritmo.}
\label{fig:tiempo_algoritmos}
\end{figure}

Mediante el análisis de tiempo empleado en cada una de las operaciones del programa mencionadas anteriormente, se determinaron las partes del algoritmo que presentaban mayores demoras.
Como se puede ver en la Fig.\ref{fig:tiempo_algoritmos}, las operaciones más costosas desde el punto de vista computacional son la de filtrado de la nube de puntos, llamada \texttt{filter} y la operación de marcado/liberación, llamada \texttt{ClearFrustums}.
Esta última comprende el acceso a las celdas del mapa global, utilizando los métodos provistos por la biblioteca OpenVDB.



Estas nubes de puntos pueden llegar a ser muy densas, del orden de millones de puntos, y al ser utilizadas en el marcado de celdas para el mapa global, es necesario realizar una compresión de las mismas.
En \cite{stvl2020steve} esto es realizado mediante un filtro de \textit{voxelizado}, disponible en la biblioteca de PCL.
Este filtro recibe como entrada nube de puntos, la cual tiene que estar en el sistema de referencia global, para poder limitar la altura de la nube de puntos.
Este límite en el eje $\mathbf{z}$, permite al \texttt{navigation\_stack} de ROS \cite{tbd} proyectar la nube de puntos sobre el plano horizontal y generar un mapa de ocupación, utilizado en la navegación del robot.
El filtro disponible en la biblioteca PCL es calculado en la CPU, y según el conocimiento de los autores, no hay otra implementación que pueda ser ejecutada en esta plataforma.

Por otro lado, los algoritmos implementados en CUDA para el filtrado de nubes de puntos normalmente se implementan en GPUs de escritorio con gran capacidad de memoria.
Esto los convierten en poco atractivos para utilizar en plataformas embebidas debido a su limitada capacidad de memoria, generalmente compartida con la CPU.

Se implementó una versión modificada de \textit{Cubic Subspaces - Neighboring Buckets} \cite{bkedkowski2013general}, implementada en CUDA y utilizando el proceso \textit{Zero-Copy} disponible en plataformas como la Jetson Nano. 
La idea principal es usar la GPU para descomponer el espacio 3D en una cuadrícula regular de $2^n \times 2^n\times 2^n$ cubos $(n = 4,5,6,7,8,9)$. Por lo tanto, para cada punto, se consideran solo 27 cubos $(3^3)$ para encontrar los vecinos más cercanos.
Para calcular la distancia entre dos puntos $p1={x_1,y_1,z_1}$ y $p2={x_2,y_2,z_2}$ se utiliza la distancia euclídea definida como: %
{
\setlength{\abovedisplayskip}{1pt}
\setlength{\belowdisplayskip}{15pt}
\begin{equation}
d(p1,p2) = \Big[ (x_2-x_1)^2+(y_2-y_1)^2+(z_2-z_1)^2\Big]^{\dfrac{1}{2}}
\end{equation}%
}
con $x$, $y$, $z$ y $d(p1,p2)$ pertenecientes al espacio $\mathbb{R}^3$.
Cada punto del espacio tridimensional $XYZ$ es normalizado, de tal forma que $\left\{x,y,z\in \mathbb{R}: -1 \leq x,y,z  \leq 1\right\}$.
Luego, son clasificados mediante un árbol de decisión para determinar a que subdivisión del espacio $2^n \times 2^n\times 2^n$ pertenecen.

La cantidad de puntos que pertenecen a cada subdivisión, es determinada mediante una tabla (\texttt{tabla\_subdiv}) ordenada de pares ``clave-valor"\,utilizando el conocido algoritmo ``Radix Sort".
Es importante notar que esta tabla es almacenada en la memoria global de la GPU, por lo tanto, todos los hilos de CUDA pueden acceder a los datos.
El par ``clave-valor", junto con la información de la cantidad de puntos en cada subdivisión, son accesibles mediante la memoria de la GPU, y se utilizará para buscar el vecino más cercano en el algoritmo.

\begin{figure*}[ht]
  \includegraphics[width=\textwidth,height=4cm]{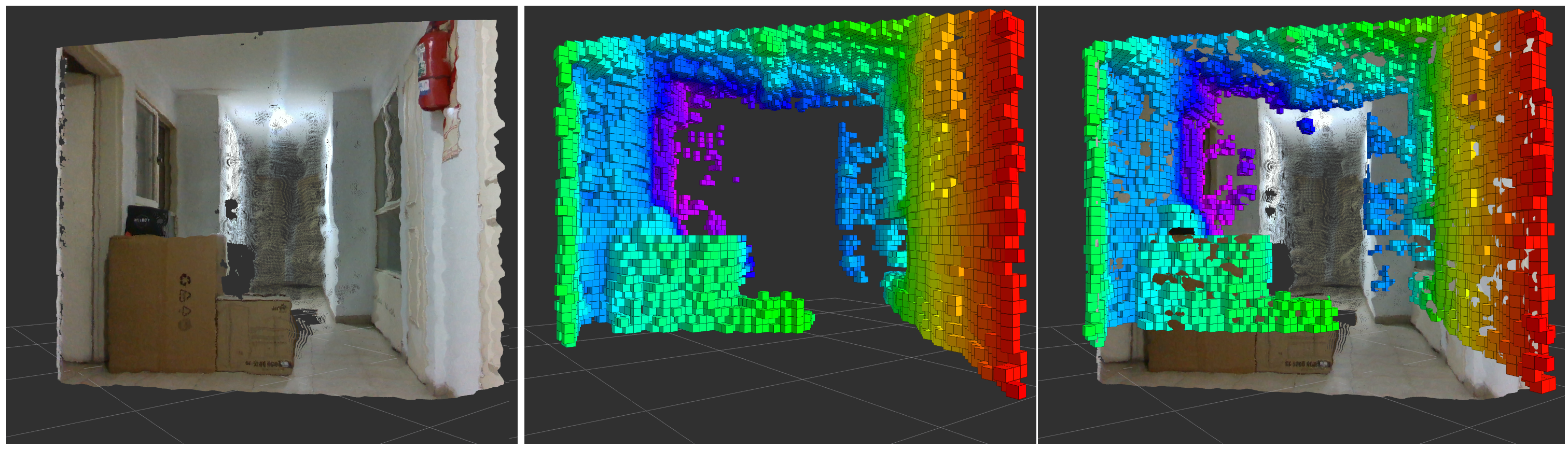}
  \caption{Secuencia de imágenes del ambiente utilizado para las pruebas. Izquierda: Nube de puntos generada por la cámara RGB-D, Centro: Grilla de ocupación 3D generada por el algoritmo STVL, Derecha: Nube de puntos y mapa superpuestos. }
  \label{fig:grilla_3d_generada}
\end{figure*}

\begin{algorithm}[ht]
\caption{Filtrado de puntos 3D}
\label{alg:filtrado}
\begin{algorithmic}[1]
\renewcommand{\algorithmicrequire}{\textbf{Entrada:}}
\renewcommand{\algorithmicensure}{\textbf{Salida:}}
\renewcommand{\algorithmicfor}{\textbf{para}}
\renewcommand{\algorithmicdo}{\textbf{hacer}}
\renewcommand{\algorithmicend}{\textbf{fin}}
\renewcommand{\algorithmicwhile}{\textbf{mientras}}
\renewcommand{\algorithmicif}{\textbf{si}}
\renewcommand{\algorithmicthen}{\textbf{entonces}}

\REQUIRE Puntero a la nube de puntos de la cámara
\ENSURE  Puntero a la de puntos filtrada

\STATE copiar el puntero del host al device
\STATE llamado a la función de CUDA
\FOR {todos los puntos $m^i_{xyz}$ en paralelo}
\STATE buscar $subdiv_{m^i}$
\STATE actualizar \texttt{tabla\_subdiv}
\ENDFOR
\STATE en paralelo ordenar \texttt{tabla\_subdiv} $\{$radix sort$\}$

\WHILE {la cantidad de puntos marcados $>$ 1000 $\{$un kernel CUDA por cada punto $m_{xyz}$ $\}$}
\FOR {todos los puntos $m^i_{xyz}$ en paralelo}
\STATE buscar $subdiv$
\FOR {todas las $subdiv$ vecinas}
\STATE contar la cantidad de vecinos de $\{$teniendo en cuenta los puntos marcados para borrar$\}$
\ENDFOR $\{$un kernel de CUDA por un punto$\}$
\STATE marcar $m^i_{xyz}$ para borrar si \texttt{cont} $>$ \texttt{umbral}
\ENDFOR $\{$un kernel de CUDA para todos los puntos$\}$
\IF {cantidad de puntos marcada $>$ 1000 }
\STATE aleatoriamente elegir 1000 puntos marcados para eliminarlos definitivamente
\ENDIF
\ENDWHILE

\STATE sincronizar el llamado a los kernels
\STATE eliminar los puntos marcados
\STATE copiar el puntero del device al host
\end{algorithmic}
\end{algorithm}

El objetivo del filtrado es eliminar puntos, para reducir la densidad de la nube de puntos y al mismo tiempo eliminar el ruido de la medición proveniente de la cámara de profundidad.
Una vez calculados los vecinos más cercanos por el método anteriormente descripto, las subdivisiones del espacio son iterativamente reducidas hasta obtener la densidad de puntos deseada.
Este proceso puede ser descripto mediante el Algoritmo \ref{alg:filtrado}. 

La nube de puntos obtenida del filtro, es una versión comprimida de la nube de entrada, y la densidad de la misma depende del tamaño de bloque elegido para el filtro.
Esta dimensión puede ser configurada al inicio del algoritmo, y también es utilizada por la estructura de OpenVDB para almacenar los puntos en el mapa global.
 
Debido a las restricciones impuestas por la evasión de obstáculos, es una prioridad garantizar cálculos completos de la nube de puntos en el menor tiempo posible. 

Para asegurar la utilización completa de la GPU, el programa determina la cantidad máxima de procesadores CUDA disponibles, para distribuir cada \texttt{subdiv} del espacio.
Esto se puede ver en el punto número 11 del Algoritmo \ref{alg:filtrado}.

Se realizaron experimentos en ambientes simulados y reales para validar el método de filtrado.
A continuación se presenta una breve descripción del sistema utilizado.

\begin{figure}[b]
\centering
  \includegraphics[width=0.45\textwidth]{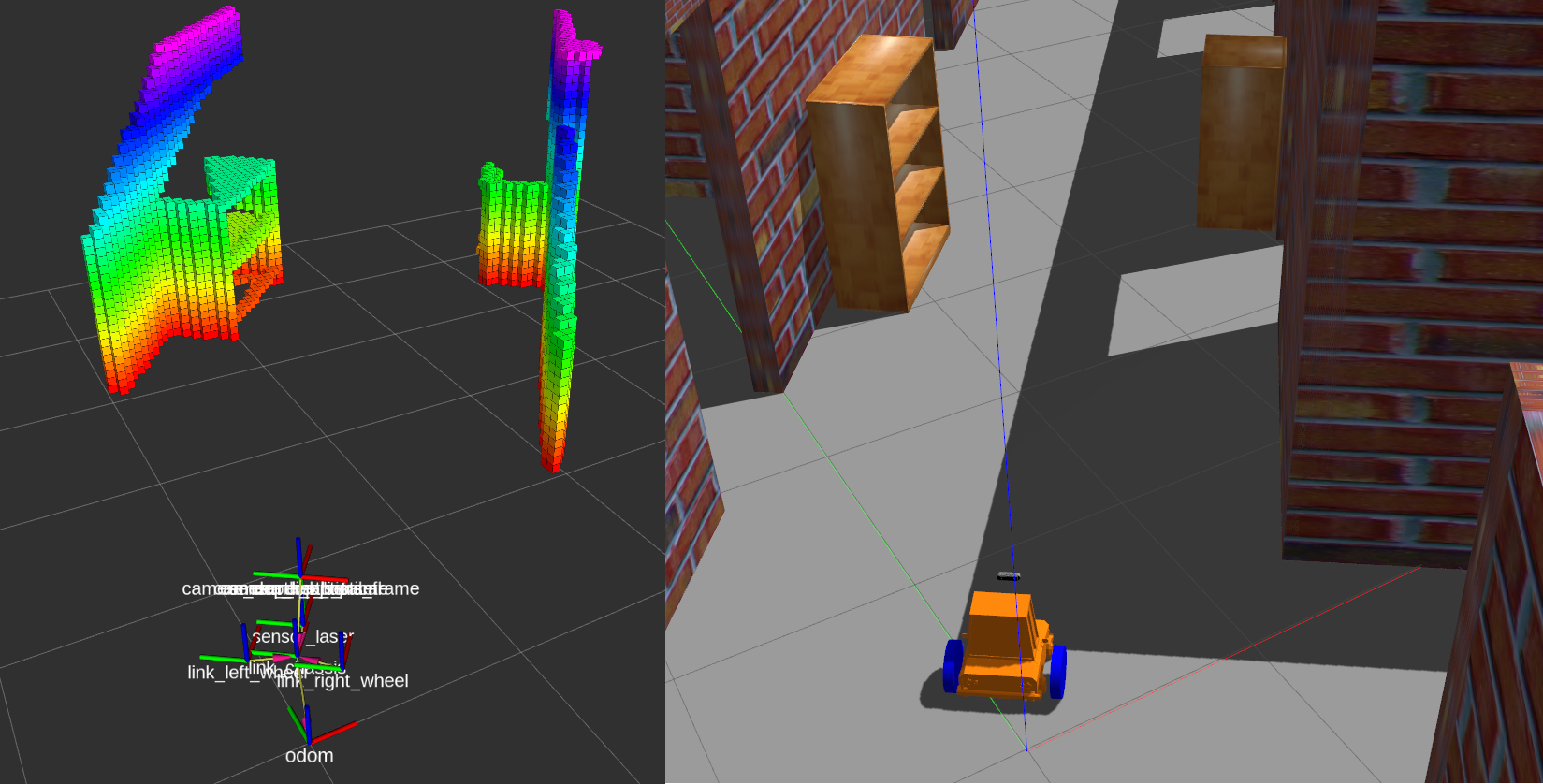}
  \caption{Robot en el entorno de simulación. Izquierda: Mapa 3D obtenido por el algoritmo. Derecha: Pasillo de oficina con obstáculos en sus laterales}
  \label{fig:prueba_simulada}
\end{figure}

\subsection{Escenario simulado}
Para evaluar el algoritmo propuesto, en una primera instancia se utilizó el simulador Gazebo \cite{koenig2004design}, en su versión 9.
Se optó por por este simulador ya que es compatible con el protocolo de mensajes de ROS Melodic utilizada en este trabajo, y cuenta con soporte para los sensores requeridos por el algoritmo.

El robot es del estilo tracción diferencial, con una cámara RGB-D montada en su parte superior.
El mismo cuenta con odometría, con lo cual el mapa generado puede ser extendido más allá del alcance de la cámara de profundidad.
En la Fig.\ref{fig:prueba_simulada}, se puede observar el mapa generado por el algoritmo.
El mismo intenta recrear un ambiente de oficinas, en el cual se encuentran dispuestos mobiliario, en ambos lados. 
Es importante destacar que al ser un sensor simulado, no presenta los artefactos normalmente encontrados en cámaras RGB-D.

\subsection{Escenario real}
Para un análisis cualitativo se realizaron pruebas en un escenario real en donde se tomaron imágenes de un pasillo con distintos obstáculos y puertas.
Se utilizó la configuración mostrada en la Fig.~\ref{fig:pruebas_real}.
La misma está compuesta por la cámara RGB-D con el eje óptico alineado con el sentido de circulación del pasillo.
La cámara fue montada sobre un perfil de aluminio.
Debajo de la misma, se colocó la plataforma de desarrollo Jetson Nano.

\section{Resultados}

\begin{figure}[b]
\centerline{\includegraphics[width=0.5\textwidth]{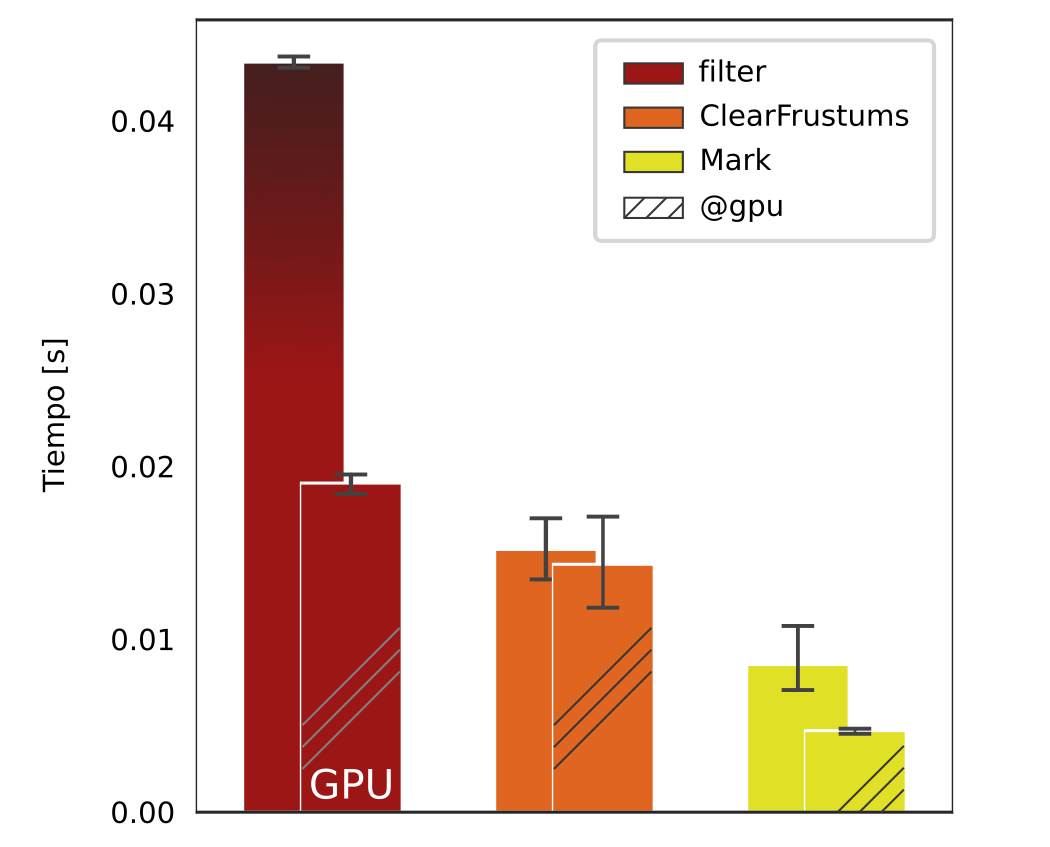}}
\caption{Comparación de las funciones de mayor carga computacional, en el algoritmo con el filtrado en CUDA y en PCL. Se puede apreciar como las funciones las funciones \texttt{clear\_frustum} y \texttt{mark} que están implementadas en CPU, ahora se ejecutan en un menor tiempo, ya que la CPU dispone de más recursos.}
\label{fig:pcl_gpu}
\end{figure}

Como se explicó anteriormente, el mayor cómputo del algoritmo está concentrado en dos partes principales, el filtrado y el acceso a la nube de puntos global mediante las funciones provistas por OVDB. 
Luego de la implementación del algoritmo, en la Fig.~\ref{fig:pcl_gpu} se muestra el análisis de cada función comparando la versión original con la versión en GPU ejecutándose en la Jetson Nano.

\begin{figure}[ht!]
\centerline{\includegraphics[width=0.5\textwidth]{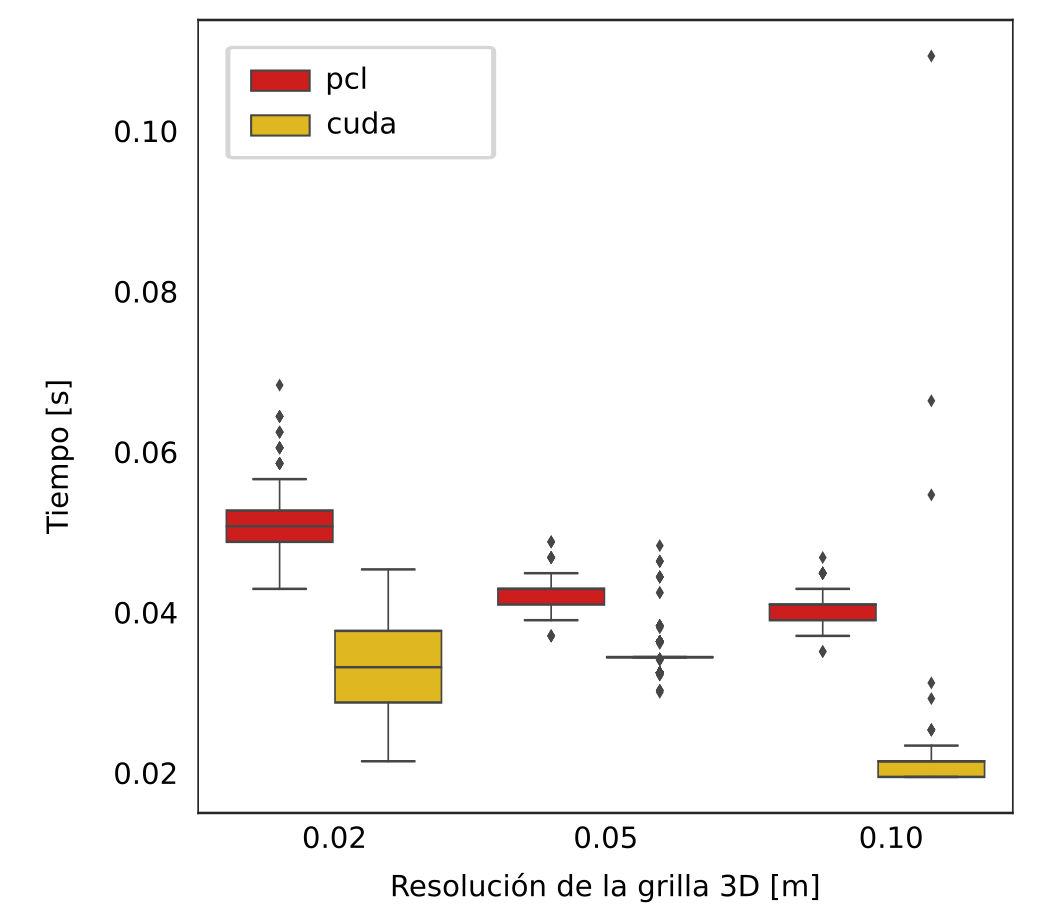}}
\caption{Comparación entre la versión original del algoritmo de filtrado mediante PCL y la versión propuesta en CUDA, para diferentes resoluciones del mapa global}
\label{fig:tiempo_filtrado}
\end{figure}

En la Fig.~\ref{fig:tiempo_filtrado}, se pueden observar los tiempos empleados por el filtrado de la nube de puntos.
Como se puede ver, la versión utilizando CUDA es más rápida que la versión original utilizando el filtrado con la biblioteca PCL.
Esto era de esperar, ya que el filtrado puede ser descompuesto de forma tal, que se expone el paralelismo del algoritmo.
Estas tareas masivamente paralelas, hacen que las placas GPU saquen grandes ventajas frente a su versión serializada.

\begin{figure}[b]
\centerline{\includegraphics[width=0.5\textwidth]{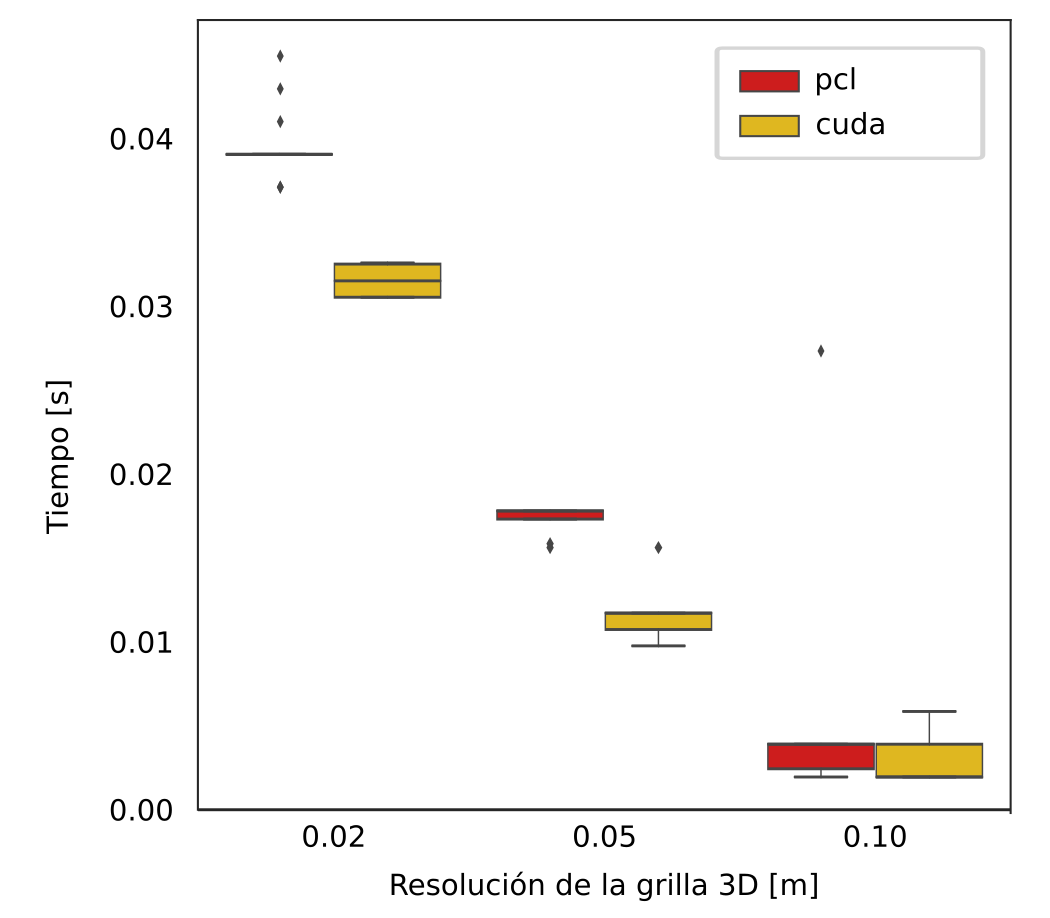}}
\caption{Comparación del tiempo para las operaciones en la CPU por parte de la biblioteca OpenVDB en el mapa global, para diferentes tamaños de grilla.}
\label{fig:tiempo_operador_ovdb}
\end{figure}

Por otro lado, llevar el procesamiento de la nube de puntos a la GPU, implica que la CPU ahora dispone de más recursos, dejando lugar para ejecutar otros algoritmos en forma simultanea.
Esto se puede ver en la Fig.~\ref{fig:tiempo_operador_ovdb} como una reducción en el tiempo de procesamiento de la nube de puntos por parte del algoritmo de OVDB, el cual se ejecuta en la CPU.
Para el mismo tamaño de grilla elegido, al algoritmo se procesa en menor tiempo en la versión con el filtro paralelizado en CUDA.

Esto también se puede ver en la Fig.~\ref{fig:pcl_gpu}, en la cual las funciones \texttt{clear\_frustum} y \texttt{mark} que están implementadas en CPU, ahora se ejecutan en un menor tiempo, pese a que no se realizaron mejoras en esas funciones.
Es importante destacar que la mejora se realizó sobre el filtrado de la nube de puntos; y el algoritmo de inserción y eliminación de puntos, mediantes las funciones de OVDB en el mapa global no presenta modificaciones.

En la Fig.~\ref{fig:grilla_3d_generada} podemos observar el mapa de ocupación 3D generado por el algoritmo.
También se puede apreciar en la parte inferior, la ausencia de voxels (puntos en el mapa 3D), debido a que es una condición requerida por el \texttt{navigation\_stack} para realizar la proyección sobre el plano y generar el mapa de ocupación.
Este mapa generado, puede ser utilizado por algoritmos de planificación, para desplazarse en el ambiente evitando obstáculos, y de forma paralela para generar un mapa con la información de odometría.
Este escenario sería el caso de un robot navegando por un pasillo, en el cual se encuentran presentes obstáculos que obstruyen su camino.
En este caso la resolución elegida fue de 5cm, la cual no solo permitió representar con efectividad los obstáculos (Cajas de cartón en la imagen), si no que también permitió capturar detalles del ambiente.
Por ejemplo, en la esquina superior derecha, podemos observar como el matafuegos es capturado por el mapa 3D.
En el caso de la puerta a la izquierda de la escena, la discontinuidad se puede distinguir por un cambio en el color de la grilla 3D.
Es importante remarcar que, debido a la presencia de ruido en la cámara de profundidad, en la tercera imagen de la Fig.~\ref{fig:grilla_3d_generada}, algunos voxels del mapa 3D son ocluidos por la nube de puntos.

El algoritmo de filtrado implementado en CUDA no tiene una gran carga aritmética, debido a que la mayoría de operaciones corresponden a movimientos de memoria.
Como se mencionó anteriormente, la plataforma utilizada dispone de una memoria LPDDR4 con cuatro canales de 16 bits, capaz de alcanzar un ancho de banda máximo teórico de 25.6GB/s, sin embargo en las pruebas realizadas la velocidad no supera los 16GB/s para copias entre la misma GPU y de 10GB/s para copias CPU-GPU.
Las pruebas fueron realizadas con los ejemplos sugeridos por el fabricante \cite{cudaSpeed}, para evaluar el ancho de banda.
Es importante destacar que las transferencias CPU-GPU se deben a no utilizar memoria fijada (\texttt{pinned\_memory)}), la cual evita que el sistema operativo la mueva o la cambie al disco.
La memoria fija proporciona una mayor velocidad de transferencia para la GPU y permite la copia asincrónica

\begin{figure}[htbp]
\centerline{\includegraphics[width=0.5\textwidth]{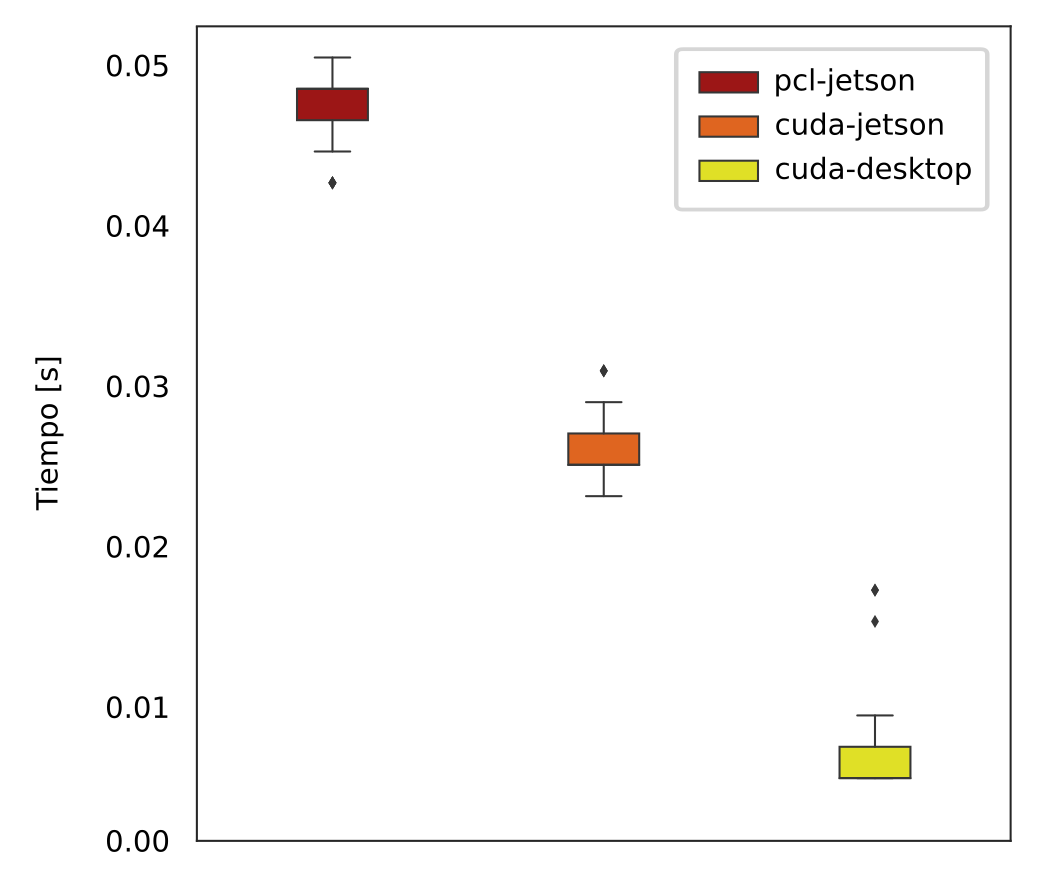}}
\caption{Comparación de tiempo de filtrado para tres configuraciones con una resolución de 0.02m: a la izquierda, el algoritmo de filtrado original corriendo en la CPU de la Jetson; al centro, el algoritmo propuesto corriendo en la GPU de la Jetson; a la derecha, el algoritmo propuesto corriendo en la GPU de escritorio}
\label{fig:mejora_velocidad}
\end{figure}

Como puede observarse en la Fig.~\ref{fig:mejora_velocidad}, se obtiene una mejora notable en el tiempo de filtrado de la nube de puntos.
La velocidad del filtrado se reduce en un orden de magnitud cuando el algoritmo es ejecutado en la computadora de escritorio como se indica en la Tabla \ref{tab:mejora_velocidad}.
Esto también se debe al cambio de micro arquitectura, ya que en Turing los procesadores se rediseñaron para unificar la memoria compartida, el almacenamiento en caché de texturas y el almacenamiento en caché de carga de memoria, en una sola unidad.
Esto se traduce en 2 veces más ancho de banda y más de 2 veces más capacidad disponible para la caché L1 \cite{cudaTuring}, en comparación a la arquitectura anterior a Turing.
Por otra parte, la placa de video cuenta con 22 SM (Streaming Multiprocessors) con con 64 CUDA cores cada uno, lo que mejora el desempeño del algoritmo, puesto que cada subdivisión del espacio (\texttt{subdiv}) es procesada en forma independiente del resto. 

Cabe aclarar que el algoritmo para este último caso, está utilizando memoria unificada, pero no puede utilizarse el concepto de \textit{Zero-copy} disponible en la plataforma Jetson.

\begin{table}[htbp]
\caption{Datos estadísticos obtenidos para la función de filtrado en la plataforma Jetson Nano y en una GPU discreta, para un tamaño de 0.02m}
\label{tab:mejora_velocidad}
\centering
\begin{tabular}{l|c|c}
Plataforma & Media [ms]& Desviación estándar\\
\hline
pcl-jetson& $46.88$ & $1.60\cdot 10^{-3}$\\
cuda-jetson& $23.2$ & $1.38\cdot 10^{-3}$\\
cuda-desktop& $3.9$ & $1.76\cdot 10^{-3}$\\
\end{tabular}
\end{table}

\balance

\section{Conclusiones}
En este trabajo se presentó una mejora sobre sobre el filtrado de una nube de puntos, en el programa STVL.
Se realizaron pruebas simuladas como en la vida real, y se analizaron las ventajas de procesar la nube de puntos en la GPU embebida en plataformas como las NVIDIA Jetson.
También se realizó una comparación de velocidad con una placa de video discreta en una PC de escritorio, en la cual se pudo comprobar que el aumento en la velocidad de memoria disminuye el tiempo de procesamiento del algoritmo.
Durante el desarrollo del algoritmo, se pudo constatar que es muy importante tener la misma arquitectura de GPU que la de la plataforma sobre la cual se está desarrollando.
Una discordancia en la misma, podría provocar errores conceptuales y algorítmicos, debido a que algunas funciones pueden no estar disponibles.
Para el caso de la Jetson Nano, la misma cuenta con una versión limitada de las funciones de memoria unificada.
Es trabajo futuro implementar mejoras sobre el algoritmo presentado en la plataforma Jetson TX2, la cual cuenta con micro arquitectura Pascal y mejor soporte de memoria unificada, con lo cual se espera mejorar aún más el rendimiento.

También se está trabajando en reemplazar el uso de las funciones de OpenVDB (realizadas en CPU), por las de la biblioteca GVDB provistas por NVIDIA \cite{gvdb}.
Debido al uso de funciones implementadas a partir de la micro arquitectura Pascal, se planea utilizar la plataforma Jetson TX2.


\bibliographystyle{IEEEtran}
\bibliography{IEEEabrv,bib/mybib.bib}

\vspace{12pt}

\end{document}